\documentclass[conference]{IEEEtran}
\IEEEoverridecommandlockouts
\usepackage{cite}
\usepackage{amsmath,amssymb,amsfonts}
\usepackage{algorithmic}
\usepackage{graphicx}
\usepackage{textcomp}
\usepackage{xcolor}
\usepackage{verbatim}
\def\BibTeX{{\rm B\kern-.05em{\sc i\kern-.025em b}\kern-.08em
    T\kern-.1667em\lower.7ex\hbox{E}\kern-.125emX}}

\IEEEoverridecommandlockouts\IEEEpubid{\makebox[\columnwidth]{ 979-8-3503-3439-5/23 /\$31.00~\copyright~2023 IEEE \hfill} \hspace{\columnsep}\makebox[\columnwidth]{ }}
    
\begin{document}

\title{Economical Quaternion Extraction from a Human Skeletal Pose Estimate using 2-D Cameras\\

\thanks{The authors are with the Center for Internet of Things, Department of Computer Science and Engineering, PESU Ring Road Campus (Bangalore, India). \\
Our code repository can be found at : https://github.com/SR42-dev/human-pose-quaternion-extraction}
}

\author{

\IEEEauthorblockN{Sriram Radhakrishna}
\IEEEauthorblockA{\textit{Department of Computer Science and Engineering} \\
\textit{P.E.S University, Banashankari}\\
Bangalore, India \\
sriram.radhakrishna42@gmail.com}

\and

\IEEEauthorblockN{Adithya Balasubramanyam}
\IEEEauthorblockA{\textit{Department of Computer Science and Engineering} \\
\textit{P.E.S University, Banashankari}\\
Bangalore, India \\
adithyab@pes.edu}

}

\maketitle

\begin{abstract}
In this paper, we present a novel algorithm to extract a quaternion from a two dimensional camera frame for estimating a contained human skeletal pose. The problem of pose estimation is usually tackled through the usage of stereo cameras and inertial measurement units for obtaining depth and euclidean distance for measurement of points in 3D space. However, the usage of these devices comes with a high signal processing latency as well as a significant monetary cost. By making use of MediaPipe, a framework for building perception pipelines for human pose estimation, the proposed algorithm extracts a quaternion from a 2-D frame capturing an image of a human object at a sub-fifty millisecond latency while also being capable of deployment at edges with a single camera frame and a generally low computational resource availability, especially for use cases involving last-minute detection and reaction by autonomous robots and similarly modelled intelligent transport systems. The algorithm seeks to bypass the funding barrier and improve accessibility for robotics researchers involved in designing control systems. 
\end{abstract}

\begin{IEEEkeywords}
quaternions, 2-D camera, pose estimation, MediaPipe, low-power, low-latency, embedded computer vision.
\end{IEEEkeywords}

\section{Introduction}
Essential scene-analysis tasks such as pedestrian detection and localization generally involve the generation of a quaternion to estimate the orientation of a target object. At times, these techniques involve the usage of stereo cameras and inertial measurement units to match feature points \cite{fathian2017quaternion} or other such methods involving expensive hardware components with a relatively large latency and computational resource utilization. 

Intuitively speaking, generalizing the potential use cases, keeping the computation closer to the system and eliminating the expensive hardware involved in the deployed algorithm would be the way forward to tackle the cost and latency issues. In order to do this, the hardware reliant inputs being accepted by existing pose estimation models must be taken note of. For this particular use case of human pose estimation, the depth of the points located within a targeted pose object and the simultaneous threading of two image streams at minimum stands out. In order to address these issues, a system containing a single 2-D camera with a human subject in frame on the hardware side to minimize costs was opted for. Additionally, a specialized deep learning based solution on the software side was implemented to account for the loss of depth perception as well as maintain latency on edge devices with a lower resource availability. The goal is to mimic the quaternion outputs provided by modern inertial measurement units for any given edge between adjacent joints on the human skeletal pose \cite{jouybari2017experimental} \\

\begin{figure}[htbp]
\centerline{\includegraphics[scale=0.42]{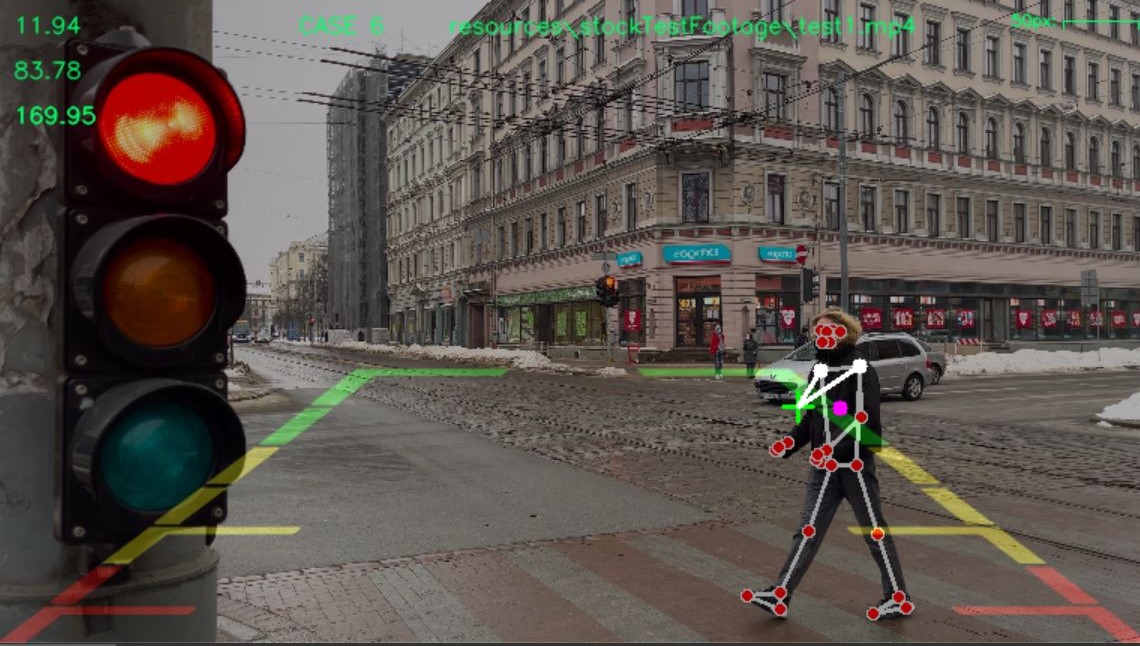}}
\caption{A demonstration of the novel algorithm for the use case of calculating the angle of orientation of a pedestrian object from a generated quaternion.}
\label{fig1}
\end{figure}

\section{Related work}
In their work on developing a quaternion-based recurrent model for human motion, Pavllo et al. noted that human motion is a stochastic sequential process with high intrinsic uncertainty \cite{pavllo2018quaternet}. While deep learning approaches have been successful in predicting human skeleton poses, they are generally computationally costly, especially for mobile robotics and autonomous vehicle navigation where cost factors are compromised in favor of larger core counts on the main edge processing unit \cite{fragkiadaki2015recurrent}. Recent drops in the retail costs of Nvidia GPU-based edge computing units have made this trade-off more viable, but the gap is best bridged by adapting CPU-based systems to more efficiently handle the necessary calculations \cite{schneider2017deeper}. Eberly noted that rotation matrices and quaternions require over 61\% fewer calculations than angle-axis representations of vector rotating operations \cite{eberly2002rotation}.

To expand on the review, recent works on body orientation estimation have been focused on improving the accuracy and robustness of the estimation algorithms. For example, the MeBoW system proposed by We et al. \cite{wu2020mebow} uses a combination of 2D pose estimation and monocular depth estimation to estimate 3D body poses and orientations accurately. Zhou et al. \cite{zhou2022joint} proposed a joint regression method that uses 2D joint positions and depth maps to estimate body orientation accurately.

In conclusion, while quaternion-based methods offer advantages over angle-axis representations, recent works have focused on improving accuracy and robustness of body orientation estimation algorithms. Additionally, advancements in hardware technologies have made it possible to perform complex computations efficiently on the edge devices. Therefore, future works in body orientation estimation should consider these advancements to improve the performance and efficiency of the algorithms.

\section{Concept Theory and Implementation Methodology}

\subsection{Overview}
The challenges faced in this approach lie in the implementation of algorithmic solutions to finding the inputs mentioned in the introductory paragraph. First, the problem of calculating the orientation of the human pose object came with the issue of localization of body landmarks with acceptable standards of latency. Following this, was the conception of a mathematical function that extracted the quaternion from the pose data of these points. Essentially, the novelty of the algorithm lies in its ability to derive a pose quaternion from a two dimensional image of a human being in a picture.

To solve the problem of estimating the orientation, we employed MediaPipe, a framework by Alphabet Inc. that provides deployed solutions from deep learning models trained on data for human pose object detection. This package is aimed at edge devices with a low resource utilization.\cite{lugaresi2019mediapipe}. While the framework achieves admirable detection latencies, an inherent flaw in the system when porting to use cases requiring quaternion generation like robotic navigation \cite{sarabandi2019survey} and such due to its basis in 3-D Cartesian space.
An added advantage provided by the novel algorithm is that portability of the model is maintained even when implemented on dynamic frames of reference, e.g.; on a mobile robot in motion.

It must be noted at this point that just a single scalar value, like the orientation is not being considered to expand the scope of the tool. For autonomous vehicles, no one would consider such a detailed skeletal tracker, but rather a deep object detector that also yields orientation of objects from an RGB image. On the other hand, if a human-robot collaboration scenario was in question, then having detailed skeletal tracker and human pose prediction is a valuable tool.

\subsection{Quaternion Transformations}
Before we get to the equations that constitute the novel algorithm, it is important to gain some clarity on why a quaternion is necessary for this use case. We can visualize this by taking the example of a pedestrian obstruction to an autonomous vehicle or robot. For situations requiring a quick response from the robot to a major last-minute change in scene, especially ones which could endanger human lives such as humans crossing the path of a moving vehicle  

To provide some context here, the quaternions can be calculated for any adjacent pairs of joints in the human pose, due to which we have narrowed down a use case for the algorithm to the problem of pedestrian orientation detection and modeled our implementation based on that. Hence, the two points chosen to evaluate the pose of our target human are the shoulder points as their relative positions are an accurate descriptor of the orientation of said target \cite{rungruangbaiyok2021shoulder}. The same equations and thought processes can be used to extract this data for any pair of joints.

\begin{figure}[htbp]
\centerline{\includegraphics[scale=0.40]{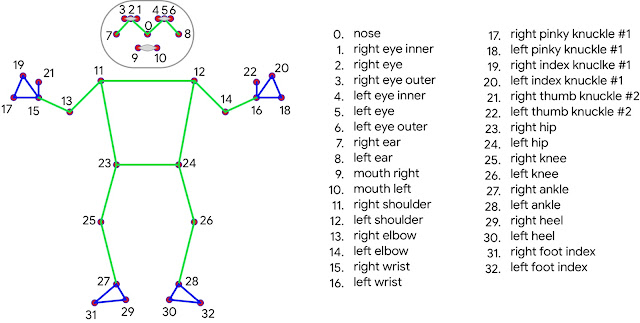}}
\caption{BlazePose 33 keypoint topology for a human skeleton as COCO (colored with green) superset \cite{bazarevsky_grishchenko_2020}}
\label{fig2}
\end{figure}

To narrow down this choice, all the provided options in the MediaPipe pose solutions documentation were evaluated \cite{bazarevsky_grishchenko_2020} for appropriate pairs of points whose projections are sure to change location on a 2-D frame when the target's roll, pitch or yaw changes. Additionally, it was also arbitrarily determined that symmetry must be maintained for these points across the mid-line of the body as this seemed like the intuitive guideline to maintain given the potential use cases of our implementation.

After obtaining the coordinates for these points from the Cartesian space estimated by the framework, a rotation matrix for the skeleton was generated by taking the coordinates of the two shoulder points mentioned before. Let us denote the points for the left and right shoulder points on the frame as (xl, yl, zl) and (xr, yr, zr) respectively. Therefore, the Z-axis vector of the rotation matrix can be calculated using the difference between the two points as 

\begin{equation}\label{eq1}
\begin{split}
        \vec{z} =  \begin{bmatrix} xl &yl  &zl \end{bmatrix} - \begin{bmatrix} xr &yr  &zr \end{bmatrix} \\
  \end{split}
\end{equation}

\begin{equation}\label{eq2}
\begin{split}
        \widehat{z} = \left\{\begin{matrix}\frac{\vec{z}  }{|\vec{z}|}; \!\ \Delta z \neq  0 
\\ 
\begin{bmatrix}
0 &-1  &0 
\end{bmatrix}; \!\ \Delta z =  0 \\
\end{matrix}\right.
  \end{split}
\end{equation}

Similarly, the X and Y axis vectors are generated as such -

\begin{equation}\label{eq4}
\begin{split}
        \widehat{x} = \begin{bmatrix}
0 &0  &1 
\end{bmatrix} \!\ \times \!\ \widehat{z} \!\ ; \!\ iff \!\ \widehat{x} \neq 0 \!\ else \begin{bmatrix}
1 &0  &0 
\end{bmatrix}
  \end{split}
\end{equation}

\begin{equation}\label{eq5}
\begin{split}
        \widehat{y} = \widehat{z} \!\ \times \!\ \widehat{x}
  \end{split}
\end{equation}

... which allows us to derive our rotation matrix \cite{hamilton1866elements} using the camera look-at method \cite{prunier_2016} as

\begin{equation}\label{eq6}
\begin{split}
        R_{3,3} = \begin{bmatrix}
r_{00} &r_{01}  &r_{02} \\ 
r_{10} &r_{11}  &r_{12} \\ 
r_{20} &r_{21}  &r_{22}
\end{bmatrix} = \begin{bmatrix}
\widehat{x} &\widehat{y}  &\widehat{z} 
\end{bmatrix}
\end{split}
\end{equation}

Subsequently, the quaternion \textbf{Q} was obtained in a standard form from this rotation matrix \cite{hamilton1866elements} as 

\begin{equation}\label{eq7}
\begin{split}
        \textbf{Q} = a + bi + cj + dk
\end{split}
\end{equation}

... where the coefficient values can be mapped according to equations 7 through 10 \cite{shuster1993survey}. 
\begin{equation}\label{eq8}
\begin{split}
        a = \frac{1}{2} \sqrt{|1 + r_{00} + r_{11} + r_{22}|} 
\end{split}
\end{equation}

\begin{equation}\label{eq9}
\begin{split}
        b = \frac{r_{21} - r_{12}}{4a}
\end{split}
\end{equation}

\begin{equation}\label{eq10}
\begin{split}
        c = \frac{r_{02} - r_{20}}{4a} 
\end{split}
\end{equation}

\begin{equation}\label{eq11}
\begin{split}
        d = \frac{r_{10} - r_{01}}{4a}
\end{split}
\end{equation}

In order to extract a usable real world statistic from these equations and to test the practicality of the novel algorithm, the direction being faced by the human pose object in the camera frame was extracted by applying certain transformations to the obtained quaternions (hereby referred to as the angle of orientation of the human pose object). Let it be noted at this point that the quaternion values returned do in fact contain some noise due to implicit measurement uncertainties in xl, yl, zl, xr, yr and zr (refer to equation 1) as illustrated in the section IV, sub-section D. These uncertainties were compensated for with the implementation of a 1-D Kalman filter for the angle of orientation of the human pose object as it was empirically the most accurate approach found post-testing.\\

This was done by extracting the angle of rotation from the angle-axis form of the quaternion, as the axis vector itself was implicitly made to be oriented upwards from the head of the human object in the frame when the rotation matrix was extracted from the model. \\

Assuming our quaternion to be of the form ... 
\begin{equation}\label{eq12}
\begin{split}
        \textbf{Q} = cos\theta + sin\theta(xi + yj + zk)
\end{split}
\end{equation}

Theta was extracted as ...
\begin{equation}\label{eq13}
\begin{split}
        \theta = \arccos (\frac{a}{\sqrt{a^{2} + b^{2} + c^{2} + d^{2}}})
\end{split}
\end{equation}

Theta was then transformed to obtain the angle of orientation in our frame of reference such that the  axis of reference was the horizontal across the camera feed window.

\begin{equation}\label{eq14}
\begin{split}
        \theta = \frac{\theta \cdot \frac{180^{2}}{\pi}}{45} - 180
\end{split}
\end{equation}

The transformation applied here takes care of the conversion from radians to degrees as well as the fact that an x degree rotation in the quaternion translates to a rotation of 2x in a real life scenario. The equation was arrived at after collecting the values of theta for a 180 degree rotation of the human object and mapping them to the real world angle being faced by the same. The angle being calculated was essentially the one included by a perpendicular from the line connecting the shoulder points to the horizontal of the camera frame. This scene can be visualized as shown in figure 3, not taking into account the fact that the camera and the shoulder points don't exist at the same height.\\

\begin{figure}[htbp]
\centerline{\includegraphics[scale=0.4]{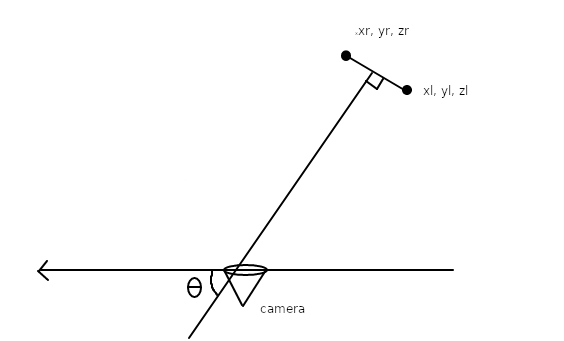}}
\caption{A simplistic visualization of the use case to detect the angle of orientation of a pedestrian object taking just the shoulder points.}
\label{fig6}
\end{figure}

\subsection{Algorithm}
Prior to covering the main algorithm, do note that the \textit{getRotationMatrix} function accepts 6 floating point values denoting xl, yl, zl, xr, yr and zr  respectively (Refer to equation 1) and returns the rotation matrix formulated in equation 5 by iterating through the calculations in equations 2 through 4. The \textit{calculateQuaternion} function accepts a rotation matrix R as given in equation 5, calculates \textbf{Q} as in equation 6 by calculating its components according to equations 7 through 10 and returns the required quaternion. \\

 The aggregated algorithmic flow for quaternion generation from the 2-D image of the skeletal pose was framed to proceed as follows -
 
 \begin{algorithmic}[1]
 \renewcommand{\algorithmicrequire}{\textbf{Input:}}
 \renewcommand{\algorithmicensure}{\textbf{Output:}}
 \REQUIRE Video feed API v
 \ENSURE  \textbf{Q}
 \\ \textit{Initialisation} :
  \STATE v = VideoCapture Class \cite{baggio2012mastering}
 \\ \textit{LOOP Process} :
  \WHILE{True}
  \STATE img = v.frame // assigning an image requested from the API to the variable img at the time of request
  \STATE pose = poseDetector(img) // detects the existence of a skeletal pose in the frame
  \STATE // getting the requested landmarks from the MediaPipe framework
  \IF {pose != None}
  \STATE xl, yl, zl = pose.getLandmarks('Left Shoulder')
  \STATE xr, yr, zr = pose.getLandmarks('Right Shoulder')
  \STATE R = getRotationMatrix(xl, yl, zl, xr, yr, zr) \\
  \STATE \textbf{Q} = calculateQuaternion(R) \\
  \RETURN \textbf{Q}
  \ENDIF
  \ENDWHILE
 \end{algorithmic}


\section{Results and Supporting Statistics}
The system was tested using a standard wide-angle USB 2.0 camera (specifications elaborated on in sub-section C) and yielded a frame rate of 24 per second on average. Taking the reciprocal of the frame rate gives us the testing latency of the algorithm as a whole. This value came out to be 41.67 milliseconds. A comparison of this method with even a rudimentary artificial neural network with three hidden layers of twenty neurons with respect to the number of mathematical operations required to generate a prediction \cite{wang2003artificial} illustrates the advantage of such a set up, given that neural networks are the current industry standard for such tasks \cite{dollar2011pedestrian} \\

A snapshot of the execution sequence can be seen in figure 4 where \textbf{Q} was obtained as 0.63 - 0.12i + 0.31j + 0.62k, as visualized in figure 4. The accuracy of the same is verified in sub-section B 'Model Accuracy Verification'. All results obtained in this section were taken from the 25th test iteration of the novel algorithm after they were deemed satisfactory.

The presented algorithm for extracting the orientation of a human from a single monocular camera appears to be efficient with a latency of 41.67 milliseconds and a frame rate of 24 per second. However, without further readings from a laser-based scanner or a stereo-camera running the appropriate image processing algorithms, it is challenging to gauge the accuracy without converting the quaternion to a human readable quantity such as angle of orientation. As such, the angle of orientation with respect to the horizon. 

Extracting this value and comparing it with the ground truth, the algorithm was able to accurately determine the angle between the line of sight of the human subject in the frame and the horizon with a constant error of 5 degrees either way. Similar results were obtained on taking the angle between the perpendicular superimposed on the midline of the subject's body and the vertical of the frame, confirming the accuracy of the system in three dimensions. This algorithm was tested on use cases that covered every possible direction of motion on a flat plane, as detailed under sub-section A. The footage recorded to contain these use cases were captured with the author as the subject on-site and solely constituted the dataset that the algorithm was tested on\\

Regarding the comparison of mathematical operations required for prediction, it is difficult to determine the relevance of the comparison outside of the use case of this algorithm's implementation on intelligent transport systems. However, several recent works have proposed deep neural network-based approaches for the same task with promising results. For instance, Yan et al. proposed a deep neural network-based method for estimating the human body pose and orientation from a monocular image with an average latency of 100 milliseconds and an accuracy of 92.8\% on a benchmark dataset \cite{yan2018spatial}. Similarly, Mehta et al. proposed an end-to-end deep learning framework for estimating 3D pose and shape from a single image, achieving state-of-the-art performance on several benchmark datasets \cite{mehta2017monocular}.

\begin{figure}[htbp]
\centerline{\includegraphics[scale=0.3]{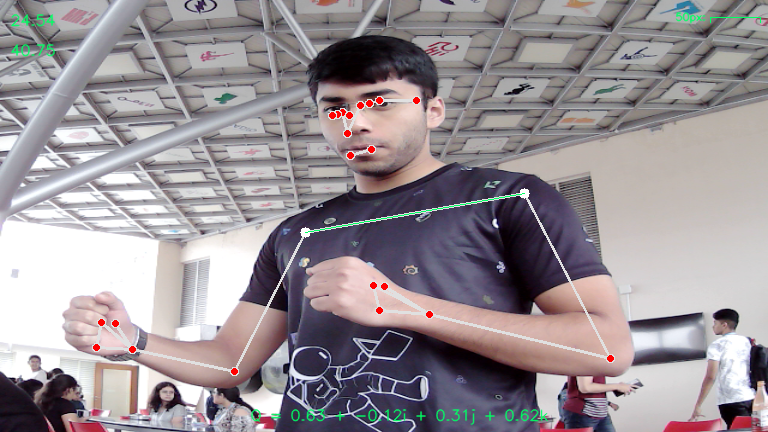}}
\caption{A demonstration of the quaternion extraction from the pose using the novel algorithm.}
\label{fig3}
\end{figure}

\subsection{Testing Conditions}
The scenes employed in the testing of the novel algorithm were intentionally chosen to reflect the systems intended use case according to the authors, i.e.; estimating the angle of orientation of a pedestrian object from the point of view of an autonomous robot or vehicle. As such, they were made to contain multiple human objects to demonstrate the arbitrary selection of the skeletal frame with the highest confidence value due to the singularly threaded nature of the novel algorithm. \\

The angle domain of 10 to 180 degrees was maintained through cases where the pedestrain was facing both towards and away from the camera by extrapolating the direction of motion of the pedestrian object on the frame and classifying them into fuzzy states for the same.

\subsection{Model Accuracy Verification}

Due to current resource limitations, the angle of orientation of the human pose object was extracted to verify the accuracy of the quaternion model by cross-checking the angle being faced by the user of the program as illustrated by equations 11 through 13. 

Note that only the pose object itself and the shoulder landmarks need to be detected for all calculations following the same in the algorithm.

\subsection{System specifications}
This system was tested on a platform with the following specifications - 
\begin{itemize}[
    \setlength{\IEEElabelindent}{\dimexpr-\labelwidth-\labelsep}
    \setlength{\itemindent}{\dimexpr\labelwidth+\labelsep}
    \setlength{\listparindent}{\parindent}
  ]
  \item Intel Core i5 7200U processor 
  \item 8GB RAM
  \item 512 GiB Solid State Drive
  \item External USB 2.0 30 FPS 2MP 'Passport' camera with a resolution of 1920x1080 and view angle of 110 degrees.
\end{itemize}

\subsection{Kalman filter to eliminate quaternion noise}
In order to obtain the angles of orientation quoted in sub-section B, a Kalman filter in one dimension was applied to the theta readings in order to get a stable feed \cite{huang2009application}. \\

To re-iterate, the angle measurements were transformed as given in equation 13 to account for the fact that all values for theta were taken with respect to an axis horizontally bisecting the frame, which was effectively achieved by taking test cases of human pose objects with erect postures. Refer to figure 6 for a visualization of the same.

\begin{figure}[htbp]
\centerline{\includegraphics[scale=0.55]{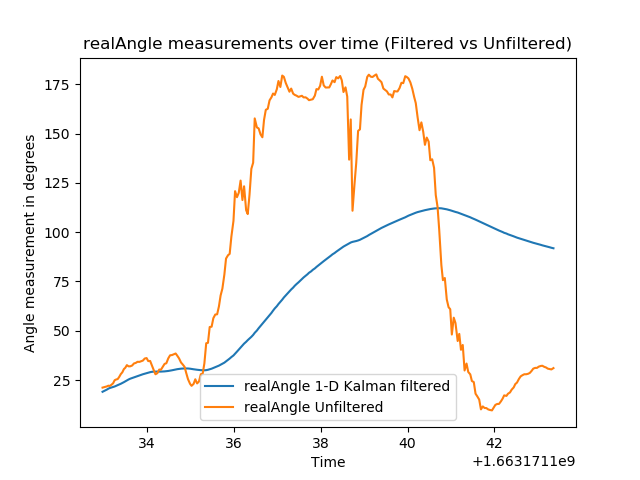}}
\caption{A line graph representation of the filtered values from the theta measurement, referred to here as realAngle.}
\label{fig11}
\end{figure}

The filter designed was made to be dynamic for a system assumed to be perfect, i.e.; the predicted covariance equation assumed the model uncertainty to be zero. This was because of the random nature of the system and no viable physical equations present to accurately define the motion of the pose object. Hence, the filter was made to accept the average of the last 10 readings in the window as the prediction when the real measurement was deemed to be not viable. For scale, 24 measurements were calculated from the generated quaternion every second. \\

Equations 14 through 16 cover the updation of the Kalman gain, state and covariance respectively, where \textit{k} is the Kalman gain, \textit{p} is the covariance, \textit{r} was the angle measurement uncertainty (emperically determined as 0.5 after testing), \textit{x} was previous measurement and \textit{z} was the current measurement itself -
\begin{equation}\label{eq15}
\begin{split}
        k = \frac{p}{p+r} \!\ ; \!\ kalman \!\ gain \!\ calculation
\end{split}
\end{equation}

\begin{equation}\label{eq16}
\begin{split}
        x = x + k(z-x) \!\ ; \!\ state \!\ updation
\end{split}
\end{equation}

\begin{equation}\label{eq17}
\begin{split}
        p = (1-k)p \!\ ; \!\ covariance \!\ updation
\end{split}
\end{equation}

In order to demonstrate the translation of this system into use cases in the real world, we chose to build a preliminary prototype of a pedestrian intent classification system \cite{kwak2017pedestrian}, where the co-ordinates of a pedestrian object on the frame as well as take the theta output value mentioned earlier to were recorded in a window of values to project the future path of motion on to the same frame. The classification was done using a fuzzy state approach taking these inputs into account. A demonstration of this is illustrated in figure 6, where one should take note of the third value in the top-right of the overlay reading 172.04. This was the angle of orientation of the pedestrian object returned by the novel algorithm in an on-ground test. This can be visually verified by the fact that the referenced pedestrian object was walking almost perpendicularly across the projected path of the robot. Also note that the image here has been flipped horizontally, and hence does not seem consistent with the previously mentioned frame bisecting axis definition at first glance. \\

\begin{figure}[htbp]
\centerline{\includegraphics[scale=0.43]{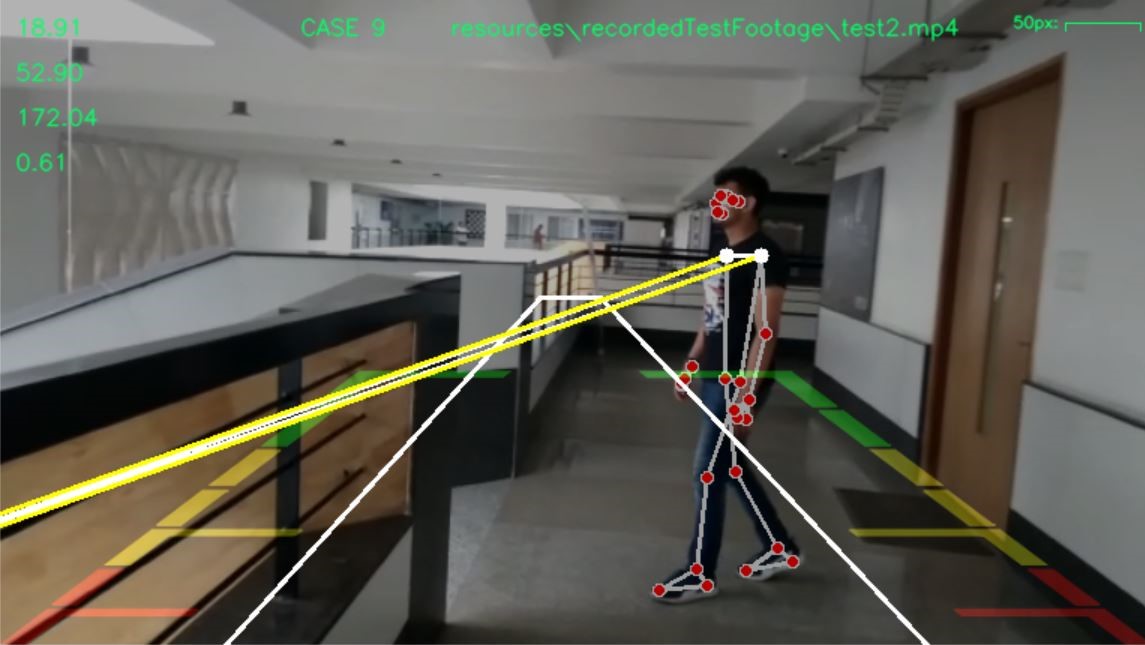}}
\caption{A demonstration of the fuzzy pedestrian intent classification algorithm prototype implemented from a ground vehicle perspective to demonstrate the practicality of quaternion extraction from a 2-D camera frame containing a human skeletal pose object.}
\label{fig12}
\end{figure}

The intention classifier was able to accurately classify approximately 75 percent of the co-ordinates of the human pose object on the frame 2 seconds into the future with a radius of accuracy of 50 pixels.

\section{Conclusions and End Notes}
Therefore, the conclusion was drawn that the novel algorithm was suitable for extracting the pose of a human object using 2-D cameras and satisfied the requirements of preserving costs due to minimal hardware usage as well as those of latency and performance as the algorithm can be implemented on hardware \cite{hegarty2014darkroom} and performs within reasonable limits of accuracy. Notwithstanding the latter, its single threaded nature as well as low computational resource usage make it suitable for edge deployment applications, although it should be noted that the number of threads required increases linearly with the number of pose objects detected.

\subsection{Optimizations and Future Scope}
 The novel algorithm can be optimized in a variety of ways, with the primary approach being multi-threading of the video feed as well as the implementation of the algorithm for detecting multiple human objects and processing their pose estimations in parallel \cite{roszyk2022adopting}. Expanding on the use case mentioned in section IV, sub-section A 'Testing Conditions', the model can be used to implement a more elaborate version of the pedestrian intent classifier demonstrated in sub-section D of the same section with more reliable prediction of whether or not a pedestrian object will cross the path in front of an autonomous robot or vehicle such that its motion may endanger the lives of the same \cite{volz2016data}. \\

In addition to optimizing the algorithm, making use of the large core count of a GPU has also shown promising prospects \cite{laguna2009comparative}. There is a growing need to develop algorithms that can fully exploit the potential of these powerful devices at this time. However, a deeper literature survey into the topic would have to be conducted to say so for sure. It is important to investigate the limitations and challenges associated with GPU computing and how they can be overcome to achieve optimal performance.

\bibliography{references.bib}
\bibliographystyle{IEEEtran}\

\end{document}